\documentclass{article}
\usepackage{iclr2027_conference,times}

\usepackage{amsmath,amsfonts,bm}

\def\eqref#1{equation~\ref{#1}}

\def\1{\bm{1}}

\DeclareMathAlphabet{\mathsfit}{\encodingdefault}{\sfdefault}{m}{sl}
\SetMathAlphabet{\mathsfit}{bold}{\encodingdefault}{\sfdefault}{bx}{n}

\usepackage{graphicx}
\usepackage{booktabs}
\usepackage{float}
\usepackage{placeins}
\floatstyle{ruled}
\newfloat{algorithm}{tbp}{loa}
\floatname{algorithm}{Algorithm}
\usepackage{hyperref}
\usepackage{url}

\title{VideoEvolve: Evolving Agent Harnesses for Video Temporal Grounding}

\author{Bingjun Luo \\ Tsinghua University \And Yuhuan Fan \\ Columbia University \And Jialin Guo \\ Harbin Engineering University \And Siqi Li \\ Tsinghua University}

\iclrfinalcopy

\begin{document}
\raggedbottom

\maketitle
\lhead{}
\renewcommand{\headrulewidth}{0pt}

\begin{abstract}
Video temporal grounding aims to localize events in videos from natural-language queries. For agents built around frozen video-language models, the harness determines how queries guide temporal predictions and how those predictions are refined. Manually refining these harnesses requires diagnosing grounding failures and coordinating changes to both agent workflows and instructions. We introduce VideoEvolve, a framework that automatically evolves agent harnesses for video temporal grounding. VideoEvolve uses a Cloze-Structured Harness Representation that preserves stage interfaces while leaving agent workflows and instructions open to evolution. Branch-Guided Harness Evolution preserves promising code branches for continued refinement, using execution feedback to guide local edits and validation to determine which improvements are carried forward. Experiments demonstrate improved grounding performance across multiple benchmarks. Component analyses identify instruction refinement as a consistent source of gains, while the benefits of evolved code vary across evaluation settings. Together, these results support automated harness evolution as an effective approach to improving video temporal grounding. Code is available at \url{https://github.com/bingjunluo/VideoEvolve}.
\end{abstract}

\section{Introduction}
\label{sec:introduction}

Video temporal grounding localizes the interval in a video that matches a natural-language query, connecting video understanding to temporally grounded retrieval and reasoning. Identifying the right event and its boundaries can remain difficult when similar actions recur or a query specifies relations between events~\citep{zhang2026timelens,zhang2026dart}. For example, the same action may be clearly recognizable at several points in a video, while its position before or after another event determines which occurrence matches the query. An agent may need to inspect additional evidence, revisit an initial prediction, or resolve competing interval hypotheses. How these decisions are organized affects the path from interpreting video content to selecting a temporal interval. For agents built around video-language models, this process depends on both the model and the execution procedure surrounding it.

\begin{figure}[t]
    \centering
    \includegraphics[width=\linewidth]{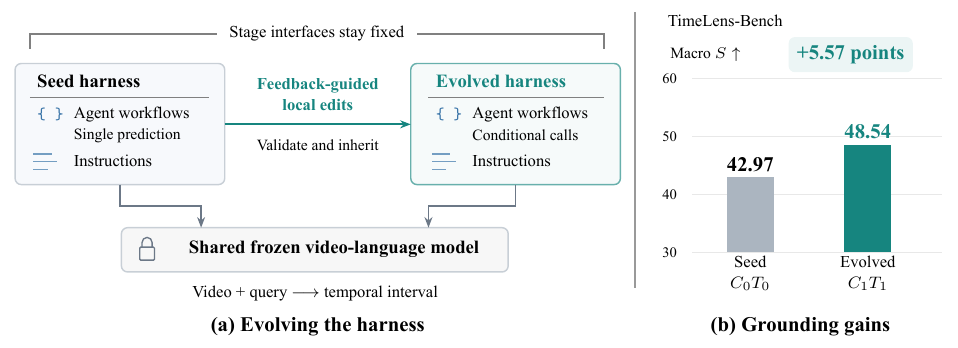}
    \caption{\textbf{VideoEvolve improves the harness around a frozen model.} (a) Workflows and instructions evolve within stable stage interfaces, guided by execution feedback and validation (schematic). (b) Evolving the harness improves macro $S$ on TimeLens-Bench (vertical axis starts at 30).}
    \label{fig:intro-teaser}
\end{figure}

Recent work improves temporal grounding through instruction tuning~\citep{ren2024timechat,huang2024vtimellm}, reinforcement learning~\citep{wang2025timer1}, and self-evolving model training~\citep{jung2026evoground}. Alongside these advances in model learning, agentic approaches organize query decomposition, grounding, verification, and adaptive reasoning into executable workflows~\citep{zheng2024tfvtg,liu2026videomind,zhang2026dart}. The workflows and instructions surrounding a model form its \emph{harness}~\citep{lee2026metaharness,zhang2026selfharness}. Improving a grounding harness requires coordinating how instructions describe an event with how code obtains and combines predictions. For example, a revised rule for distinguishing repeated events may call for changes to when a prediction is checked or which alternative interval is selected. Manually making such changes requires diagnosing failures and translating them into reusable procedures. A correction for one ambiguous query may change behavior elsewhere, so successful predictions also need to inform how the harness is revised. This motivates automatically improving the harness while keeping the target video-language model frozen.

Program-based agent design already offers a foundation for this approach: ADAS and Meta-Harness search over executable agent programs~\citep{hu2024adas,lee2026metaharness}, while the Darwin G\"odel Machine retains historical programs as starting points for further improvement~\citep{zhang2025dgm}. We organize harness evolution for temporal grounding around two requirements. First, changes to hypothesis generation must remain compatible with downstream verification and selection. The representation should therefore preserve the interfaces that carry observations, interval hypotheses, and verification evidence, while leaving internal workflows and instructions open to revision. Second, an intermediate code change may require further edits before the complete harness is ready for adoption. Search should allow such programs to continue developing, while using evaluation of the complete candidate to decide which changes are inherited. A local edit must be assessed in the context of the full harness, since its effect can depend on existing instructions and downstream decision rules. Together, these requirements connect an editable grounding representation with a search procedure that separates continued exploration from acceptance.

We introduce \textbf{VideoEvolve}, a framework that evolves grounding harnesses using labeled development feedback while keeping the target model frozen (Figure~\ref{fig:method}). Its \emph{Cloze-Structured Harness Representation} preserves stage interfaces while allowing executable workflows and instructions within them to evolve. Execution traces, grounding errors, and successful predictions from the current harness provide evidence about which decisions should be revised and which behavior should be preserved. \emph{Branch-Guided Harness Evolution} uses execution feedback to guide local edits, preserves promising code branches for continued refinement, and validates complete candidates before adopting them as the basis for subsequent iterations. We evaluate the resulting harnesses on five grounding benchmarks. Figure~\ref{fig:intro-teaser} illustrates the approach and its gain over the initial harness. VideoEvolve exceeds the compared frozen-model workflows on Charades-STA and ActivityNet, and matched comparisons show gains over its seed on all five benchmarks. Component analyses identify instruction refinement as a consistent source of gains, while code contributions vary across benchmarks. This distinction motivates evaluating both grounding quality and inference usage when assessing an evolved harness.

Our contributions are threefold:
\begin{itemize}
    \item We introduce VideoEvolve and a Cloze-Structured Harness Representation that supports evolving grounding workflows and instructions within stable stage interfaces around a frozen video-language model.
    \item We develop Branch-Guided Harness Evolution, connecting feedback-guided local edits and continued code-branch refinement with validation-based inheritance of complete harnesses. Development feedback guides branch retention, while validation determines which complete candidates are accepted and inherited by subsequent iterations.
    \item We evaluate the evolved harnesses on five grounding benchmarks, using search controls, code--instruction exchanges, and deployment-cost measurements to characterize performance, component contributions, and inference cost.
\end{itemize}

\section{Related Work}
\label{sec:related-work}

\subsection{Video Temporal Grounding}
Feature-based grounding methods learn temporal predictions from video--text representations. MH-DETR~\citep{xu2023mhdetr} and UVCOM~\citep{xiao2024uvcom} improve cross-modal interaction and temporal modeling for moment retrieval and highlight detection. Video-language models extend this line through architectures and training procedures that connect visual content to timestamps. TimeChat~\citep{ren2024timechat} introduces time-aware encoding and instruction tuning, VTimeLLM~\citep{huang2024vtimellm} develops boundary-aware training, and LLaVA-MR~\citep{lu2024llavamr} combines frame and time encoding with generative moment prediction. Time-R1~\citep{wang2025timer1} and TimeLens~\citep{zhang2026timelens} investigate reinforcement learning with verifiable rewards for temporal localization. More recently, EvoGround~\citep{jung2026evoground} constructs training signals from raw videos through a self-reinforcing proposer--solver loop that updates both models. These approaches improve learned grounding components through task-specific training. VideoEvolve uses labeled development feedback to optimize a reusable harness around a frozen video-language model, directing improvement toward its executable workflow and instructions.

\subsection{Agentic Video Understanding}
Agentic and workflow-based methods organize video evidence into intermediate decisions before producing a prediction. For temporal grounding, TFVTG~\citep{zheng2024tfvtg} decomposes queries into sub-events and combines their localization results using temporal relations, while ChatVTG~\citep{qu2024chatvtg} combines clip captioning, query matching, and moment refinement. DART~\citep{zhang2026dart} adapts the reasoning route to query difficulty. Beyond these grounding workflows, VideoTree~\citep{wang2024videotree} expands a query-relevant hierarchy of visual evidence, and VideoMind~\citep{liu2026videomind} coordinates planning, grounding, verification, and answering through role-specific LoRA adapters. These approaches illustrate how evidence selection, additional observations, and prediction refinement can be organized within an agent. Their procedures can adapt to individual videos and queries; VideoEvolve additionally treats the programs implementing such decisions as objects of development-time search. Execution feedback guides changes that persist in the harness used for subsequent grounding tasks.

\subsection{Automated Agent Design and Harness Evolution}
Automated agent design broadens optimization from model inputs to the systems that construct and act on them. ADAS~\citep{hu2024adas} represents agent designs in code, allowing search over executable workflows. Meta-Harness~\citep{lee2026metaharness} optimizes task-specific harness programs using prior source code, scores, and execution traces. Self-Harness~\citep{zhang2026selfharness} uses the evaluated model itself to diagnose weaknesses and propose bounded harness edits, which are then evaluated through regression tests. VideoEvolve follows this program-based view and structures the editable space around the information exchanged during video grounding. Its Cloze-Structured Harness Representation preserves stage interfaces while allowing internal workflows and instructions to evolve. This structure maintains the passage of observations, interval hypotheses, and verification evidence as their processing rules change; an external coding agent proposes the modifications.

Search history provides alternative starting points for further improvement. The Darwin G\"odel Machine (DGM)~\citep{zhang2025dgm} maintains an archive of agents whose code can be modified, including intermediate designs that serve as stepping stones. AutoHarness~\citep{lou2026autoharness} organizes code hypotheses in a tree and selects nodes for feedback-guided refinement. Meta-Harness retains evaluated programs and allows its proposer to inspect prior artifacts without imposing a parent-selection rule. VideoEvolve's Branch-Guided Harness Evolution organizes exploration around a current champion and a retained code branch, allowing further edits to a program that has not yet replaced the champion. Validation determines whether a complete candidate is adopted and inherited by subsequent iterations. This separates continued exploration from acceptance within the stage-structured grounding harness.

\section{VideoEvolve}
\label{sec:method}

\begin{figure}[t]
    \centering
    \includegraphics[width=\linewidth]{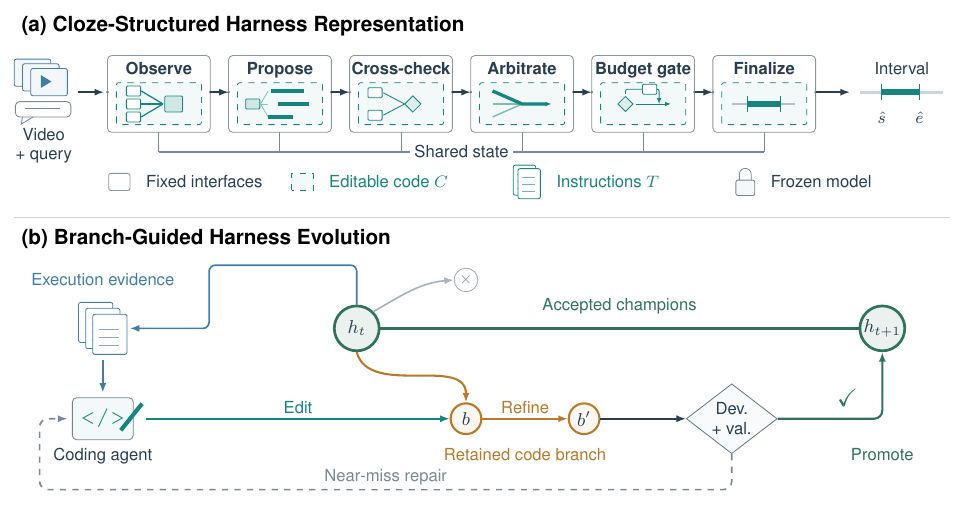}
    \caption{\textbf{VideoEvolve.} (a) Fixed stage interfaces contain editable workflows, with instructions evolving alongside code. (b) Retained code branches are refined off the accepted-harness mainline. Development comparison and validation gate promotion, while eligible validation rejections provide a repair path.}
    \label{fig:method}
\end{figure}

A video grounding harness coordinates how a model interprets a query, revisits visual evidence, and resolves competing interval predictions. VideoEvolve improves this procedure by evolving the harness around a frozen video-language model. Its \emph{Cloze-Structured Harness Representation} preserves the interfaces through which grounding stages exchange information, while leaving their workflows and instructions editable. \emph{Branch-Guided Harness Evolution} uses execution feedback to refine candidate programs, retaining intermediate code branches for further exploration. Validation determines which complete harness is inherited by subsequent iterations. Figure~\ref{fig:method} illustrates the framework.

\subsection{Task and Harness Optimization}
\label{sec:task}

Given a video $V$ of duration $d$ and a natural-language query $q$, video temporal grounding predicts an interval $\hat y=(\hat s,\hat e)$ with $0\leq\hat s\leq\hat e\leq d$. We represent a harness as $h=(C,T)$: executable code $C$ controls model calls, state updates, and interval selection, while text $T$ supplies task instructions and reusable guidance (\emph{skills}). Together, they determine the agent workflow. With a frozen target model $M$ and tool interfaces $U$, the prediction is $\hat y=\operatorname{Exec}(M,U,h;V,q)$.

Following program-based agent design and harness optimization~\citep{hu2024adas,lee2026metaharness}, we optimize $h$ while keeping $M$ and $U$ fixed. Let $\mathcal H_B$ contain executable harnesses that satisfy the stage interfaces and per-instance resource limits $B$. Starting from a working seed $h_{\mathrm{seed}}$, we seek to
\begin{equation}
    \max_{h\in\mathcal H_B} S_{\mathrm{dev}}(h;M),
    \label{eq:objective}
\end{equation}
where $S_{\mathrm{dev}}$ measures grounding performance on labeled development data. Search proceeds through local edits under the same resource limits and returns the final accepted harness.

\subsection{Cloze-Structured Harness Representation}
\label{sec:representation}

A change to how interval hypotheses are generated must remain compatible with how later stages verify and select them. We therefore separate a stable execution skeleton from editable stage implementations and instructions. The representation is \emph{cloze-structured}: designated parts of a working harness can be rewritten while the interfaces connecting them remain fixed. The editable contents are executable code and text, allowing search to introduce new decision rules within this structure.

The six stages in Figure~\ref{fig:method} organize observation, hypothesis generation, cross-checking, arbitration, additional resource use, and finalization. Shared state carries visual observations, candidate intervals, verification evidence, and the current selection between stages. Within each stage, code can revise model interactions, tool use, and hypothesis processing; instructions can revise event matching and boundary judgments. Fixed stage order thus supports flexible internal workflows. For example, the \texttt{budget\_gate} stage can obtain another prediction and revise the selected interval before finalization, allowing refinement to occur within the existing interfaces.

Each proposal modifies one permitted code or text surface and inherits the rest of its parent harness. Candidates must pass interface checks before evaluation as complete executable harnesses. Successive inherited edits can therefore combine changes to code and instructions across iterations, while each proposal targets a local modification.

\subsection{Branch-Guided Harness Evolution}
\label{sec:evolution}

An intermediate code change may need further revision before the complete harness is suitable for adoption. VideoEvolve therefore maintains two roles: the \emph{champion} contains accepted improvements and supplies the current execution evidence, while a retained \emph{code branch} provides an alternative parent for continued refinement. A branch can accumulate edits before it qualifies to replace the champion.

At iteration $t$, executing the champion $h_t$ on the current development batch produces predictions, execution traces, and ground-truth-based evaluations. We assemble these into evidence $E_t$, including failures and a weakness summary to guide revision, successful cases to preserve, and feedback from earlier attempts. An external coding agent uses this evidence and the permitted edit surface to propose candidate $j$:
\begin{equation}
    h_{t,j}=\operatorname{Edit}(p_{t,j},E_t),
    \label{eq:edit}
\end{equation}
where $p_{t,j}$ is the parent being edited. The current evidence describes the champion's behavior even when the edit starts from a different parent.

Ordinary proposals start from $h_t$. Code proposals can instead extend the retained code attempt with the highest recorded development score among those evaluated under the current champion. Retention makes an unaccepted program available for another edit; it does not change the active champion. The branch resets after a promotion or when its retry allowance is exhausted, anchoring subsequent exploration to the current accepted harness.

Validation rejections provide targeted repair feedback. A candidate that improves the development score but fails validation can become a \emph{near miss}. Eligible near misses receive a bounded repair opportunity: the proposal starts from the champion, or the retained branch for code edits, and uses the rejected candidate's changes together with aggregate regression diagnostics to guide the edit. Branch retention uses development scores, while repair responds to a specific validation rejection. Both mechanisms generate candidates for the same selection procedure.

\subsection{Validation-Guided Harness Selection}
\label{sec:selection}

Selection determines which exploratory changes become the next accepted harness. Because development batches vary across iterations, each valid candidate is compared with the champion on the same batch. Only the highest-scoring candidate that strictly improves on the champion proceeds to validation; otherwise, the champion is retained and code attempts remain eligible for branch continuation.

The candidate and champion are then compared on a validation batch separated from the development batch by video. Acceptance requires grounding performance to remain within the permitted regression limits and execution to satisfy resource and reliability constraints. A fixed reference harness also anchors the quality check. Rejection preserves the champion and provides aggregate diagnostics for possible repair. Per-example validation content is withheld from the proposer, but its aggregate feedback can influence later edits; validation is therefore part of search, separate from the final endpoint evaluation.

At most one complete candidate replaces the champion in each iteration. Subsequent proposals inherit its evaluated combination of code and instructions. Search returns the last champion, which is deployed with the frozen model and tools. Algorithm~\ref{alg:evolution} summarizes the loop.

\begin{algorithm}[!htbp]
\caption{Branch-Guided Harness Evolution}
\label{alg:evolution}
\small
\begin{tabbing}
\hspace{1em}\=\hspace{1em}\=\hspace{1em}\=\hspace{1em}\=\kill
\textbf{Input:} frozen model $M$, tools $U$, limits $B$, seed $h_{\mathrm{seed}}$,\\
\> development and validation data\\
$h\gets h_{\mathrm{seed}}$; initialize code-branch and repair state\\
\textbf{while} stopping condition is not met \textbf{do}\\
\> Execute $h$ on the current development batch; construct evidence $E$\\
\> Select parents from $h$ or the retained code branch; add eligible near misses as repair context\\
\> Propose local edits and retain candidates passing interface checks\\
\> Evaluate candidates and $h$ on the same development batch\\
\> $c\gets$ highest-scoring valid candidate, if any\\
\> \textbf{if} $c$ exists and its development score exceeds that of $h$ \textbf{then}\\
\>\> Evaluate $c$ and $h$ on the video-disjoint validation batch\\
\>\> \textbf{if} the validation gate accepts $c$ \textbf{then}\\
\>\>\> $h\gets c$\\
\>\> \textbf{else}\\
\>\>\> Record the near miss and aggregate diagnostics for possible repair\\
\> Update the retained code branch from evaluated code attempts\\
\> Reset the branch after promotion or exhausted retries\\
\textbf{return} $h$
\end{tabbing}
\end{algorithm}

\section{Experiments}
\label{sec:experiments}

We evaluate grounding performance, disentangle code and instruction contributions, and examine the search and inference costs of the resulting harnesses. We then connect retained edits to observed execution behavior.

\subsection{Experimental Setup}
\label{sec:experimental-setup}
\paragraph{Benchmarks.}
We evaluate on five benchmarks: Charades-STA (test), ActivityNet (val\_2), QVHighlights (validation), ReXTime (validation), and TimeLens-Bench~\citep{zhang2026timelens}. ActivityNet includes 17,016 evaluated queries after excluding five videos with inconsistent decoded durations; ReXTime includes all 921 validation queries. TimeLens-Bench contains 9,404 queries across Charades-TimeLens, ActivityNet-TimeLens, and QVHighlights-TimeLens. We use the queries and temporal annotations released by Zhang et al., who refined these datasets through manual annotation and cross-validation.

\paragraph{Evaluation metrics.}
We report mean temporal IoU (mIoU) and recall at IoU thresholds 0.5 and 0.7 (R@.5 and R@.7), expressed as percentages. On the original QVHighlights split, predictions are scored by their maximum IoU over all annotated windows. Following the TimeLens-Bench evaluation~\citep{zhang2026timelens}, we report recall and mIoU; we aggregate each metric equally over its three datasets. For search and component analyses, we summarize the grounding metrics as $S=0.5\,\mathrm{mIoU}+0.3\,\mathrm{R@.5}+0.2\,\mathrm{R@.7}$, averaged equally across the three datasets. Failed TimeLens queries receive zero IoU and remain in the denominator; on the original benchmarks, valid fallback intervals are scored and invalid intervals receive zero IoU.

\paragraph{Models and search configuration.}
Unless otherwise stated, our harness experiments use frozen Qwen3.6-35B-A3B-FP8 (Qwen-35B). Additional experiments use gemma-4-26B-A4B-it. A coding agent using gpt-5.6-sol proposes harness edits. Each search allocates 60 proposal slots over 20 rounds, subject to per-query limits of four solver calls and 96 observed frames.

\paragraph{Comparisons.}
Published training-based methods and frozen-model workflows provide task-level references under their reported settings. Controlled comparisons hold the target model fixed: we compare the evolved harness with its exact Cloze seed and exchange seed and evolved components. These evaluations use frozen endpoints without further optimization. A separate Full/No-Branch search pair tests branch retention while preserving champion inheritance, historical feedback, validation, and repair.

Each search comparison uses one completed run per arm; the reported results compare the resulting frozen endpoints.

\subsection{Main Results on Video Temporal Grounding}
\label{sec:main-results}
\begin{table}[!htbp]
\centering
\caption{Video temporal grounding results on five benchmarks, comparing VideoEvolve with training-based and agentic methods. R@.5 and mIoU are reported as percentages; higher is better. Bold indicates the best result in each column.}
\label{tab:public-results}
\begingroup
\footnotesize
\setlength{\tabcolsep}{2pt}
\renewcommand{\arraystretch}{1.10}
\begin{tabular*}{\linewidth}{@{\extracolsep{\fill}}l*{10}{r}@{}}
\toprule
& \multicolumn{2}{c}{\shortstack{Charades-STA}} & \multicolumn{2}{c}{\shortstack{ActivityNet}} & \multicolumn{2}{c}{\shortstack{QVHighlights}} & \multicolumn{2}{c}{\shortstack{ReXTime}}  & \multicolumn{2}{c}{\shortstack{TimeLens-Bench}} \\
\cmidrule(lr){2-3}\cmidrule(lr){4-5}\cmidrule(lr){6-7}\cmidrule(lr){8-9}\cmidrule(lr){10-11}
Method & R@.5 & mIoU & R@.5 & mIoU & R@.5 & mIoU & R@.5 & mIoU  & R@.5 & mIoU \\
\midrule
\multicolumn{11}{l}{\textit{Training-based Methods}} \\
MH-DETR & 56.37 & -- & \textbf{47.15} & -- & 60.84 & -- & -- & --  & -- & -- \\
TimeChat & 32.20 & -- & -- & -- & -- & -- & 7.61 & 11.65  & -- & -- \\
VTimeLLM & 27.50 & 31.20 & 27.80 & 30.40 & -- & -- & 17.41 & 20.14  & -- & -- \\
UVCOM & 59.25 & -- & -- & -- & 65.10 & -- & -- & --  & -- & -- \\
Time-R1-7B & 60.80 & -- & 39.00 & -- & -- & -- & -- & --  & 38.17 & 39.63 \\
EvoGround & 60.50 & 53.10 & -- & -- & -- & -- & -- & --  & -- & -- \\
VideoMind-7B & 59.10 & 50.20 & 30.30 & 33.30 & -- & -- & 25.52 & 27.61  & -- & -- \\
\midrule
\multicolumn{11}{l}{\textit{Agentic and Workflow-based Methods}} \\
ChatVTG & 33.01 & 34.87 & -- & -- & -- & -- & -- & --  & -- & -- \\
TFVTG & 49.97 & 44.51 & 27.02 & 34.10 & -- & -- & -- & --  & -- & -- \\
DART & 52.04 & 48.93 & 32.14 & 39.89 & -- & -- & -- & --  & -- & -- \\
\addlinespace[2pt]
\textbf{VideoEvolve} & \textbf{70.05} & \textbf{61.35} & 40.97 & \textbf{43.28} & \textbf{71.68} & \textbf{66.27} & \textbf{59.17} & \textbf{53.93} & \textbf{53.53} & \textbf{50.73} \\
\bottomrule
\end{tabular*}
\endgroup
\end{table}

Table~\ref{tab:public-results} places VideoEvolve in the broader grounding landscape. VideoEvolve surpasses the listed frozen-model workflows on Charades-STA and ActivityNet and compares favorably with the listed training-based methods across the reported benchmarks, although MH-DETR retains higher ActivityNet recall. It also exceeds the published Time-R1 reference on both TimeLens macro metrics. Strong performance in both recall and mIoU indicates that the evolved harness improves not only the frequency of predictions above the overlap threshold but also average temporal alignment. Together, these results show that harness design offers a practical route to stronger temporal grounding without parameter updates.

Matched seed-to-evolved comparisons provide direct evidence for improvement within the same model. Relative to the exact Cloze seed, VideoEvolve improves mIoU on all five benchmarks. The gains on Charades-STA and QVHighlights extend beyond videos used for harness development and selection. Because the solver is unchanged in these comparisons, the improvements are attributable to the evolved harness as a whole, including its instructions and inference procedure. The following analyses examine branch retention, the resulting harness components, and computational cost.

\subsection{Ablation Studies}
\label{sec:ablations}
\paragraph{Contribution of branch retention.}
A separate Full/No-Branch search pair tests whether continuing promising non-champion branches helps harness discovery. Both arms start from the same seed and use the same development sampling schedule and 60 proposal slots. No-Branch removes branch continuation while retaining champion inheritance, historical feedback, validation, and repair. As Table~\ref{tab:branch-ablation} shows, branch retention improves mIoU on Charades-STA and QVHighlights and increases TimeLens macro $S$ by 1.58 points. ReXTime also improves, while ActivityNet mIoU is slightly lower. The gains on the two video-disjoint benchmarks support keeping alternatives available beyond the current champion. Full also uses fewer development-time solver calls than No-Branch, so the advantage in this pair does not come from greater solver usage.
\begin{table}[!htbp]
\centering
\caption{Branch-retention ablation; endpoints come from a separate search pair.}
\label{tab:branch-ablation}
\begingroup\footnotesize\setlength{\tabcolsep}{2pt}\renewcommand{\arraystretch}{1.10}
\begin{tabular*}{\linewidth}{@{\extracolsep{\fill}}l*{10}{r}@{}}
\toprule
& \multicolumn{2}{c}{Charades-STA} & \multicolumn{2}{c}{ActivityNet} & \multicolumn{2}{c}{QVHighlights} & \multicolumn{2}{c}{ReXTime} & \multicolumn{2}{c}{TimeLens-Bench} \\
\cmidrule(lr){2-3}\cmidrule(lr){4-5}\cmidrule(lr){6-7}\cmidrule(lr){8-9}\cmidrule(lr){10-11}
Search variant & R@.5 & mIoU & R@.5 & mIoU & R@.5 & mIoU & R@.5 & mIoU & R@.5 & mIoU \\
\midrule
No-Branch & 66.59 & 58.82 & 41.11 & \textbf{44.24} & 67.81 & 64.22 & 56.89 & 53.38 & 51.56 & 49.64 \\
Full & \textbf{70.81} & \textbf{61.90} & \textbf{41.99} & 44.04 & \textbf{72.32} & \textbf{67.24} & \textbf{58.96} & \textbf{54.56} & \textbf{53.72} & \textbf{51.05} \\
\bottomrule
\end{tabular*}
\endgroup
\end{table}

No-Branch still receives historical feedback and can revise the champion, so its lower performance on most benchmarks suggests that remembering previous outcomes does not fully replace the ability to continue editing a retained program.

\paragraph{Analysis of evolved components.}
\label{sec:components}
\begin{table}[!htbp]
\centering
\caption{Grounding quality and per-query inference cost of harness components. TimeLens-Bench quality is macro-averaged.}
\label{tab:components}
\begingroup\small\setlength{\tabcolsep}{4pt}\renewcommand{\arraystretch}{1.10}
\begin{tabular}{@{}llrrrr@{}}
\toprule
Dataset & Harness configuration & R@.5$\uparrow$ & mIoU$\uparrow$ & Calls$\downarrow$ & Input tokens (k)$\downarrow$ \\
\midrule
ReXTime & Seed & 51.14 & 48.71 & \textbf{0.996} & \textbf{3.24} \\
 & Evolved instructions only & 55.70 & 52.00 & \textbf{0.996} & 3.97 \\
 & Evolved code only & 52.99 & 49.85 & 1.963 & 6.53 \\
 & Full VideoEvolve & \textbf{59.17} & \textbf{53.93} & 1.948 & 7.89 \\
\midrule
TimeLens-Bench & Seed & 46.86 & 45.68 & \textbf{1.000} & \textbf{3.73} \\
 & Evolved instructions only & 53.02 & 50.35 & \textbf{1.000} & 4.45 \\
 & Evolved code only & 49.73 & 47.97 & 1.776 & 6.50 \\
 & Full VideoEvolve & \textbf{53.53} & \textbf{50.73} & 1.726 & 7.59 \\
\bottomrule
\end{tabular}
\endgroup
\end{table}

We exchange seed and evolved code and instructions while keeping the target model fixed, without further optimization. Instructions comprise the grounding prompt and textual skills. On ReXTime and TimeLens-Bench, replacing either seed component with its evolved counterpart improves recall and mIoU, and the full harness achieves the highest values of both metrics (Table~\ref{tab:components}). Restoring either component to its seed version lowers the full harness's performance, showing that both evolved components contribute in these evaluations. Their effects differ across settings: the conditional mIoU gain from evolved code is larger on ReXTime than on TimeLens-Bench. These results connect improvements in query interpretation with changes to prediction refinement, with the strongest evaluated configuration retaining both evolved components.

\subsection{Search and Inference Cost}
\label{sec:cost}
\paragraph{Development-time search.}
The seed-to-evolved improvements in Table~\ref{tab:components} result from a search using 60 proposal slots and 217,767 solver calls. Search cost measures the computation used to discover the harness; deployment cost measures its per-query inference usage. The search budget includes repeated candidate evaluations that are incurred during development. Once an endpoint is selected, subsequent queries execute that fixed harness without invoking the coding agent or repeating candidate selection. This separates the upfront cost of discovering a reusable procedure from the cost paid each time it is applied.

\paragraph{Quality--cost tradeoff at inference.}
Table~\ref{tab:components} also reports deployment usage. The full harness uses more calls and input tokens than the seed and instruction-only configurations on both benchmarks, while all four configurations retain the same 64 sampled frames. Its additional computation therefore revisits existing observations. Average usage varies across benchmarks, reflecting the conditional refinement policy; the quality gains come with increased inference cost. Instructions alone improve quality without adding solver calls, although their longer text increases input-token usage. The full harness uses slightly fewer calls than the code-only configuration on both benchmarks, yet consumes more input tokens. Call counts and token volume therefore capture different aspects of deployment cost: the former reflects how often predictions are revisited, while the latter also accounts for the instructions supplied on each call.

\subsection{Evolved Harness Behaviors}
\label{sec:behaviors}
\paragraph{From generic localization to explicit evidence tests.}
The seed grounding instruction asks for the segment that best matches the question. The evolved instruction adds explicit occurrence and boundary tests: ``Sweep the full timeline in chronological order and form a candidate interval for every plausible occurrence, rather than committing to the first or most salient partial match.'' It also requires that ``immediately before the start the first required action or state must not yet hold, and immediately after the end the last required action or state must no longer hold.'' These retained instructions specify how to distinguish repeated events and check temporal boundaries.

\paragraph{From a single prediction to conditional refinement.}
The seed harness returns one grounding prediction from 64 uniformly sampled frames. The evolved harness keeps these observations but changes how predictions are obtained and combined: it requests another whole-video hypothesis when the timeline is coarsely sampled or the predicted interval occupies an unusually small or large fraction of the video. Substantial disagreement can trigger a third call, after which start and end coordinates are combined by their medians. Evolution thus changes both when to revisit a prediction and how to reconcile alternatives.

\begin{figure}[!htbp]
\centering
\includegraphics[width=\linewidth]{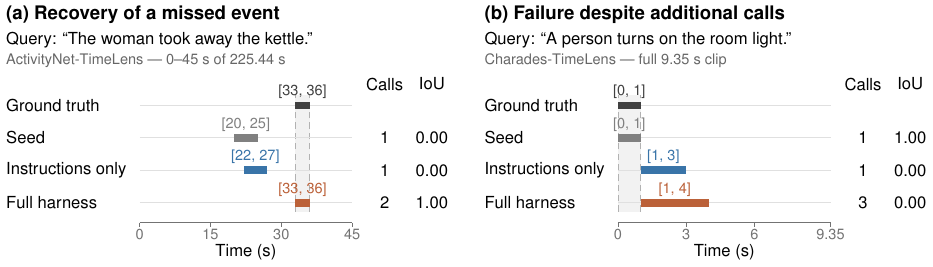}
\caption{\textbf{Grounding refinement and failure.} Final predictions from the seed ($C_0T_0$), evolved instructions alone ($C_0T_1$), and full harness ($C_1T_1$). Shading marks the ground-truth interval; calls count solver invocations.}
\label{fig:grounding-cases}
\end{figure}

\paragraph{Refinement can recover a missed interval.}
Figure~\ref{fig:grounding-cases} compares the seed, instruction-only, and full configurations on two queries. For the kettle query, the seed and instruction-only predictions precede the annotated event. The full harness executes two calls and returns the annotated interval, [33,36]\,s. In this example, replacing instructions alone does not recover the event, whereas the complete evolved procedure does. The recorded execution follows the two-hypothesis median path.

\paragraph{Refinement can still miss the event.}
For the light-switch query, the seed correctly predicts [0,1]\,s, but the instruction-only configuration shifts the interval to [1,3]\,s. The full harness uses three calls and returns [1,4]\,s, still missing the annotated event. This regression illustrates a limitation of combining model-generated hypotheses: combining additional predictions does not guarantee correct temporal alignment. Together, these cases illustrate why the aggregate benefit of executable refinement can depend on the evaluation setting: additional hypotheses can recover a missed event, but they can also preserve or amplify a localization error.

\section{Conclusion}
\label{sec:conclusion}
We presented VideoEvolve for automatically evolving video temporal grounding harnesses around frozen models. Cloze-Structured Harness Representation preserves stage interfaces while exposing workflows and instructions to revision; Branch-Guided Harness Evolution combines feedback-guided edits, continued branch exploration, and validation-based inheritance. Experiments show gains over initial harnesses across multiple benchmarks, with component analyses revealing context-dependent contributions from code and instructions. These results support harness evolution as a practical route to stronger grounding without parameter updates.

\clearpage
\subsection*{AI use statement}
In this work, we used a language-model-driven coding agent to generate and modify the executable code and task instructions of video temporal grounding agents, and used generative AI tools to assist with code editing, figure and table preparation, and reviewing the manuscript for clarity of expression. The authors reviewed the AI-assisted content and take responsibility for the final manuscript, code, and scientific claims.

\bibliography{iclr2027_conference}
\bibliographystyle{iclr2027_conference}

\end{document}